\documentclass[
]
{ceurart}

\usepackage{listings}
\usepackage{cleveref}
\usepackage{csquotes}
\usepackage{float}
\usepackage{ragged2e}

\usepackage{tikz}
\usetikzlibrary{decorations.pathmorphing}
\usetikzlibrary{decorations.pathreplacing}
\usetikzlibrary{intersections,backgrounds}
\usetikzlibrary{calc}

\usepackage[nolist]{acronym}

\newcolumntype{x}[1]{>{\centering\let\newline\\\arraybackslash\hspace{0pt}}p{#1}}
\newcolumntype{y}[1]{>{\raggedright\let\newline\\\arraybackslash\hspace{0pt}}p{#1}}

\begin{document}

\ifcasreviewlayout
    \linenumbers
\else
\fi

\copyrightyear{2024}
\copyrightclause{Copyright for this paper by its authors. Use permitted under Creative Commons License Attribution 4.0 International (CC BY 4.0).}

\conference{EWAF'24: European Workshop on Algorithmic Fairness, July 01--03, 2024, Mainz, Germany}

\title{Implications of the AI Act for Non-Discrimination Law and Algorithmic Fairness}

\author[1,2]{Luca Deck}[email=luca.deck@uni-bayreuth.de]
\address[1]{University of Bayreuth}
\address[2]{Fraunhofer FIT}
\author[1]{Jan-Laurin Müller}[email=jan-laurin.mueller@uni-bayreuth.de]
\author[1]{Conradin Braun}[email=conradin.braun@uni-bayreuth.de]
\author[1,2]{Domenique Zipperling}[email=domenique.zipperling@uni-bayreuth.de]
\author[1,2]{Niklas Kühl}[email=kuehl@uni-bayreuth.de]



\begin{abstract}
The topic of fairness in AI, as debated in the FATE (Fairness, Accountability, Transparency, and Ethics in AI) communities, has sparked meaningful discussions in the past years.
However, from a legal perspective, particularly from the perspective of European Union law, many open questions remain.
Whereas algorithmic fairness aims to mitigate structural inequalities at design-level, European non-discrimination law is tailored to individual cases of discrimination after an AI model has been deployed.
The AI Act might present a tremendous step towards bridging these two approaches by shifting non-discrimination responsibilities into the design stage of AI models.
Based on an integrative reading of the AI Act, we comment on legal as well as technical enforcement problems and propose practical implications on bias detection and bias correction in order to specify and comply with specific technical requirements.
\end{abstract}


\maketitle



\section{Introduction}

AI systems' propensity to discriminate against legally protected groups has been demonstrated across multiple social contexts, ranging from decision-support systems for criminal risk assessment \citep{Angwin.2022}, recruiting \citep{Dastin.2022}, and credit scoring \citep{FUSTER.2022}, to applications in computer vision \citep{Shankar.2017, buolamwini.2018, zhao.2017, burns.2019} and natural language processing \citep{Bolukbasi.2016, Caliskan.2017, Garg.2018}.
In light of the rapid advancements of AI, the increasing use of AI systems across multiple domains has triggered a broad and interdisciplinary debate on the ``ethics of algorithms'' \citep{Mittelstadt.2016, Kearns.2019, Martens.2022}.
Central to this debate are the FATE principles (fairness, accountability, transparency, and ethics), with fairness encompassing the social goals of non-discrimination, inclusion, and equality \citep{Floridi.2018, EuropeanCommission.HLEG.2019}.

The discourse at the interface with legal scholarship, however, is only starting to gain traction (e.g., \citep{Hacker.2018, Wachter.2021a, Wachter.2021b, HauerKevekordes.2021, Zehlike.2020, Weerts.2023}).
In this short paper, we make three contributions: First, we briefly retrace the academic discourses on non-discrimination law and algorithmic fairness to highlight their current misalignment. 
Second, we argue that the European Union's AI Act might pose a seminal link to merging these debates. 
Based on this integrative conception, we thirdly sketch how the AI Act could provide a means to solve the enforcement problems of both---non-discrimination law and algorithmic fairness---and comment on upcoming challenges regulators and developers will face when specifying and verifying technical requirements. 

\section{Non-discrimination law vs. algorithmic fairness}

\paragraph{Legal Context: Non-discrimination law and its shortcomings.}

From a legal perspective, non-discrimination law appears to be suitable to address the potential harms of unfair AI systems at first glance.
However, legal scholars from both sides of the Atlantic have demonstrated that U.S. \citep{Barocas.2016} and EU \citep{Hacker.2018} non-discrimination law alike may fall short in doing so.
One of the main deficiencies of traditional non-discrimination regimes in the context of algorithmic discrimination is law enforcement.
Enforcement has always been a central shortcoming of non-discrimination law, especially in jurisdictions that primarily rely on individual litigation (cf., \citep{Fredman.2011, Berghahn.2016}).
In such jurisdictions, individual victims face substantial problems when it comes to recognizing, proving, and bringing instances of discrimination before the courts.
AI systems exacerbate these problems \citep{Spiecker.2023}.
Due to the opacity of these systems, those affected by algorithmic discrimination are often unable to recognize instances of (potential) discrimination \citep{Burrell.2016}.
Moreover, even when individuals suspect discrimination, restricted access to models or training data severely impedes their ability to meet the requirements of the burden of proof imposed on them by procedural law \citep{Hacker.2018}.
Furthermore, European non-discrimination law is tailored to individual cases of discrimination hampering its application to broad-scale goals like designing fair AI systems.
Non-discrimination regimes, therefore, face substantial challenges when it comes to enforcing the principles of equality and non-discrimination.

\paragraph{Technical context: algorithmic fairness and its shortcomings.}

On a technical level, methods for algorithmic fairness from the field of computer science set out to fill this gap. 
By developing a plethora of technical bias definitions and fairness metrics (cf. \citep{Verma.2018, Chouldechova.2018, Pessach.2022}) as well as practical bias detection and bias mitigation techniques \citep{Hardt.2021, Deck.ACritical.2024, Hort.2023, Pagano.2023}, computer scientists try to implement ethical and legal fairness considerations ``by design'' \citep{Zliobaite.2017}.
The shortcomings of these technical fairness approaches, however, are twofold:
First, formalization and quantification will never provide answers to fundamentally normative challenges such as selecting the right fairness metric for the right context or trading off conflicting objectives \citep{Binns.2018, Friedler.2021}.
Such challenges arising from conflict between values can be supported but not be solved by formal methods \citep{Narayanan.2022}.
Second, due to its orientation towards a specific academic audience and reliance on self-governance, discourse on algorithmic fairness faces its own ``enforcement problems'' \citep{Mittelstadt.2019}.
The AI Act may alleviate both---the enforcement problems of non-discrimination law and the technical fairness discourse---alike.

\section{Implications of the AI Act}

\paragraph{Enforcement ``by design''?}

According to Recital 4a, the AI Act explicitly aims to protect the fundamental rights set out in Art. 2 of the Treaty of the European Union.
Among these rights are equality and non-discrimination in particular. 
In order to prevent algorithmic discrimination, the regulation establishes special requirements (Art. 6 et seq. AI Act) for high-risk systems in the areas of education (Recital 35), employment (Recital 36), insurance and credit (Recital 37), law enforcement (Recital 38), as well as migration (Recital 39).
However, despite its explicit goal to prevent discrimination, the regulation lacks a clear substantive standard for determining when unequal treatment is inadmissible.
According to Art. 10(2)f AI Act \textit{``[t]raining, validation and testing data sets shall be subject to data governance and management practices appropriate for the intended purpose of the AI system''} and thus have to be examined for \textit{``possible biases that are likely to [...] lead to discrimination prohibited under Union law''}.
The AI Act therefore leaves the judgment call about what constitutes illegal discrimination to existing legislation.
However, traditional non-discrimination law's requirements can only be implemented during model development (as intended by the AI Act) if they are ``translated'' into technical fairness requirements.
To achieve this goal, scholars from all domains are bound to collaborate. When doing so, they must proceed in a conscious and contextualizing manner and take into account the diverging perspectives of AI Act and non-discrimination law.
European non-discrimination law is tailored to individual instances of discrimination after an AI model has been deployed---an inherently \textit{retrospective} approach.
In contrast to this, the AI Act \textit{prospectively} demands fairness interventions by implementing non-discrimination requirements at the stage of model design.
Guidance by democratically justified institutions on how to implement such requirements might bridge the gap toward alleviating both the legal and the technical enforcement problems.

\paragraph{Enabling ``bias detection and correction''?}

Legal requirements for the development of AI systems are not only subject to the AI Act. Due to the tension between fairness and privacy during the training and evaluation stage of AI, conflicts with data protection law may equally arise.
On the one hand, ignoring personal demographic data promotes the same risk as the widely rejected idea of fairness through unawareness because legally protected attributes like race and gender usually correlate to innocuous proxy variables \citep{Kusner.2017, Dwork.2012}.
If protected attributes are unavailable during model training and evaluation, these subtle correlations cannot be accounted for, nor can technical fairness metrics be tested and optimized.
On the other hand, Art. 9 GDPR places particularly high demands on the lawful processing of personal data about special categories.
Therefore, the same sensitive data that is protected by data protection law is also essential to effectively avoid discriminatory outputs.
The AI Act seeks to mitigate this tension by broadening the scope of lawful data processing.
Art. 10(5) AI Act states that \textit{``[t]o the extent that it is strictly necessary for the purposes of ensuring bias detection and correction in relation to the high-risk AI systems [...], the providers of such systems may exceptionally process special categories of personal data referred to in Art. 9(1) [GDPR].''}
This is accompanied by Recital 44c, which adds that \textit{``[i]n order to protect the right of others from the discrimination that might result from the bias in AI systems [...] the providers should, exceptionally, [...] be able to process also special categories of personal data, as a matter of substantial public interest within the meaning of Art. 9(2)(g) [GDPR].''}
Therefore, discrimination and fairness considerations can provide a justification for data processing during the training phase of high-risk AI systems.
However, balancing the public and private interests regarding non-discrimination and privacy will inevitably lead to intricate trade-offs.



\section{Practical challenges for compliance}

\paragraph{Defining bias: what are ``appropriate'' fairness metrics?}

The discussed implications of the AI Act raise two important questions on how to put non-discrimination and fairness into practice.
First, the concept of technical fairness metrics begs the question which one(s) may be ``appropriate for the intended purpose of the AI system'' (Art. 10(2)f AI Act).
Technical fairness definitions have already been examined for their compatibility with moral norms \citep{Hellman.2020} and non-discrimination regimes \citep{Wachter.2021a, Wachter.2021b, HauerKevekordes.2021, KoutsovitiKoumeri.2023, Weerts.2023} alike. 
However, legal concepts relying on flexible ex-post standards and human intuition are in tension with the mathematical need for precision and ex-ante standardization \citep{Weerts.2023, KoutsovitiKoumeri.2023}.
Also, the interdisciplinary discourse needs to acknowledge that fairness and non-discrimination might present inherently different concepts targeted at different social contexts.
Prior works have suggested that a single standard of fairness can be achieved by ``translating'' legal non-discrimination requirements from the employment context into technical fairness metrics \citep{Wachter.2021a, HauerKevekordes.2021}.
However, the heterogeneity of social contexts (e.g., employment versus criminal sentencing) demands a corresponding flexibility in fairness requirements \citep{CorbettDavies.2018, Binns.2020}.
Instead of aiming for a one-size-fits-all solution, we therefore recommend applying the landscape of available technical fairness metrics to different legal conceptions of discrimination depending on the societal context.

\paragraph{Detecting and correcting bias: when are biases ``likely to lead to discrimination''?}

The second challenge is defining when ``possible biases that are likely to [...] lead to discrimination''.
Technical fairness metrics such as statistical parity or equalized odds offer an actionable approach to measure and mitigate ``bias'' \citep{Barocas.2019, Hardt.2021, Weerts.2023, chen2023fairness}.
However, it remains unanswered what kind of evidence would signal sufficient efforts of bias detection and correction.
Setting aside the debate on metric selection, let us assume algorithmic hiring requires male and female applicants to receive equal hiring rates (demographic parity).
Statistical hypothesis testing provides a suitable method to verify compliance with this requirement, in this case a simple z-test.
To test the hypothesis of compliance with demographic parity, we are interested in the test's error rates, i.e., falsely detecting a violation (type 1 error) or the likelihood of failing to detect a violation (type 2 error).
Notably, a larger disparity in hiring probabilities between groups and a larger sample size decreases type 2 error.
Unfortunately, the z-test is also sensitive to the acceptance rate---particularly for small sample sizes.
For example, for 1000 male and 1000 female applicants, type 2 error decreases by 0.8\% - points if only 700 instead of 900 applicants are accepted---despite identical group disparities (see \Cref{appendix}).
This effect is especially strong for imbalanced datasets. 
For 1800 male and 200 female applicants, type 2 error even decreases by 6\% - points if only 780 instead of 980 applicants are accepted---again, despite identical group disparities  (see \Cref{appendix}).
Our example highlights the need for guidance in selecting appropriate tests and specifying standards for the error rates of tests utilized in bias detection.

\section{Conclusion} \label{conclusion}
In this short paper, we outlined how the AI Act could promote the convergence of legal non-discrimination discourse and technical algorithmic fairness discourse.
While we sketch its potential implications on fairness requirements of future AI developments, specifying and enforcing concrete legal requirements will be an intricate future task.
In the absence of legal precedents, both disciplines are in need of pioneering work at the intersection of non-discrimination law and algorithmic fairness.



\bibliography{bibliography.bib}

\appendix
\section{Appendix} \label{appendix} 

\begin{figure}[h]
    \centering
    \includegraphics[width=0.98\textwidth]{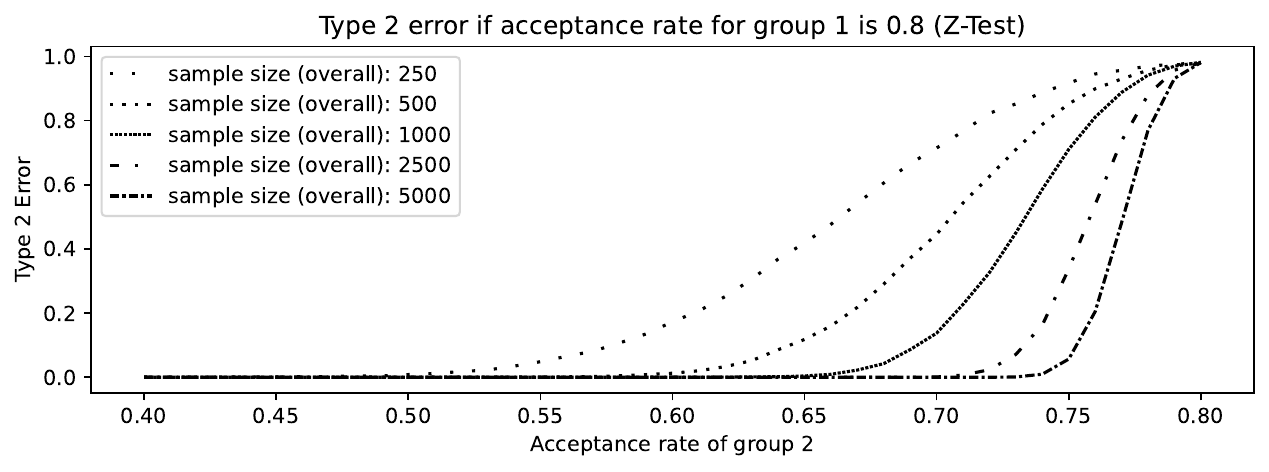}
    \caption{Type 2 error for increasing the disparity of the acceptance rate for two groups}
    \label{fig:z-zest_increases_ES}
\end{figure}

\begin{figure}[ht]
    \centering
    \includegraphics[width=0.98\textwidth]{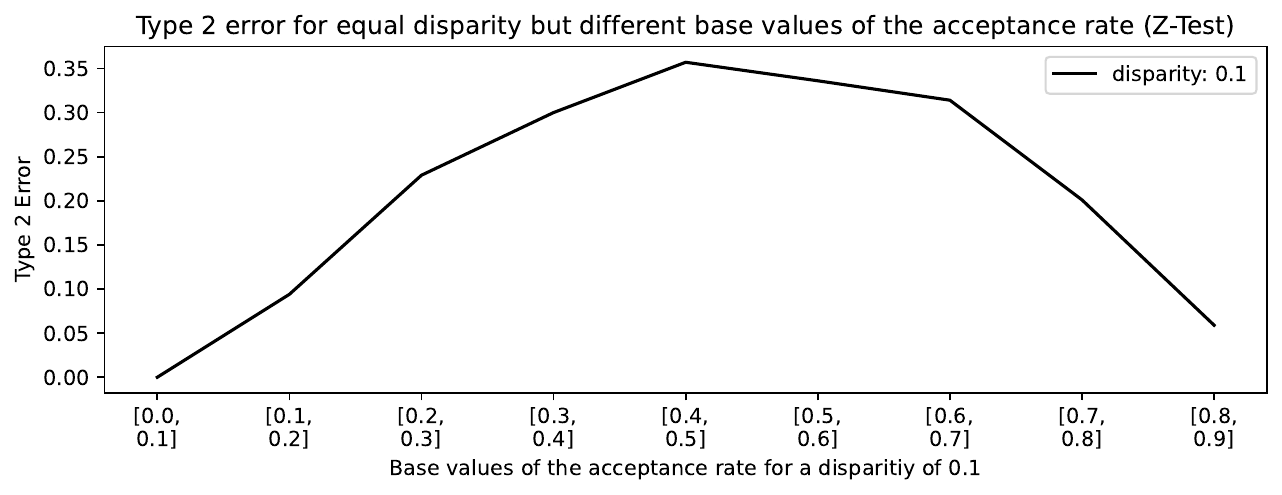}
    \caption{Type 2 error for the same disparity in acceptance rate for two groups but for different acceptance rate values}
    \label{fig:z-zest_same_ES}
\end{figure}

The appendix aims to visualize the effects described (Section 4). \Cref{fig:z-zest_increases_ES} refers to the effect of larger disparity in hiring probabilities on the probability of not detecting a violation (Type 2 error). For example, for a sample size of 2,500 a change from acceptance rate from 0.75 to 0.7 results in a 17\% - point decrease (from 33\% to 16\%) in type 2 error if group 1 has an acceptance rate of 0.8. Furthermore, it demonstrates that increasing the sample size for the same disparity also decreases the probability of a type 2 error. Doubling the sample size from 2,500 to 5,000 samples decreases the type 2 error by 27\% -points (from 33\% to 6\%). The first effect increases with increasing sample size, while the second one decreases with increasing sample size.
\Cref{fig:z-zest_same_ES} demonstrates the effect of the same disparity (0.1) but different acceptance rates. For 1800 male and 200 female applicants, the type 2 error decreases by 6\% - points if only 780 (720 male, 60 female) instead of 980 (900 male and 80 female) applicants are accepted. This effect is amplified by imbalanced data sets and small sample sizes.

\end{document}